\pdfoutput=1
\documentclass[11pt]{article}

\usepackage[final]{acl}

\usepackage{times}
\usepackage{latexsym}

\usepackage{tikz}
\usetikzlibrary{shapes.geometric, arrows, positioning}

\usepackage[T1]{fontenc}
\usepackage[utf8]{inputenc}
\usepackage{booktabs}
\usepackage{graphicx}
\usepackage{rotating}
\usepackage{listings}
\usepackage{tcolorbox}
\usepackage{xcolor}    % for color definitions
\usepackage{fvextra}

\usepackage{microtype}

\usepackage{inconsolata}

\title{Empaths at SemEval-2025 Task 11: Retrieval-Augmented Approach to Perceived Emotions Prediction}

\usepackage{inconsolata}
\usepackage{hyperref}
\usepackage{fontawesome5}

\author{ Lev Morozov\textsuperscript{\faCrow}, 
        Aleksandr Mogilevskii\textsuperscript{\faBookOpen},
     \textbf{Alexander Shirnin}\textsuperscript{\faCrow}\\
    \textsuperscript{\faCrow}HSE University,
    \textsuperscript{\faBookOpen}Independent researcher \\
    \small{
    \textbf{Correspondence:} \href{mailto:ashirnin@hse.ru}{\texttt{ashirnin@hse.ru}}
}
}

\usepackage{enumitem}

\begin{document}
\maketitle
\begin{abstract}
  
This paper describes EmoRAG, a system designed to detect perceived emotions in text for SemEval-2025 Task 11, Subtask A: Multi-label Emotion Detection. 
We focus on predicting the perceived emotions of the speaker from a given text snippet, labeling it with emotions such as joy, sadness, fear, anger, surprise, and disgust. 
Our approach does not require additional model training and only uses an ensemble of models to predict emotions. 
EmoRAG achieves results comparable to the best performing systems, while being more efficient, scalable, and easier to implement.
\end{abstract}

\section{Introduction}

SemEval-2025 Task 11~\cite{muhammad-etal-2025-semeval1} introduces a new task and a multilingual, multi-label, emotion-annotated dataset of texts. This task focuses on perceived emotion detection, aiming to determine the emotions that most readers would infer a speaker is experiencing based on a given text snippet. It does not concern the emotions evoked in the reader or the speaker’s true emotions. Instead, it addresses how emotions are commonly interpreted, recognizing that perception may be influenced by cultural context, individual expression differences, and the nuances of text-based communication. The shared task consists of three subtasks: (A) \textbf{Multi-label Emotion Detection}, that involves predicting which emotions are perceived in the speaker’s words; (B) \textbf{Emotion Intensity Prediction}, which quantifies the strength of an expressed emotion on an ordinal scale; and (C) \textbf{Cross-lingual Emotion Detection}, which assesses how well models generalize perceived emotion detection across languages using training data from a single language.

% The SemEval 2025 Task 11 focuses on bridging the gap in text-based emotion detection and offers three distinct subtasks:

% \begin{itemize}
%     \item \textbf{Track A: Multi-label Emotion Detection} \\
%     This task involves predicting the perceived emotions of the speaker from a given text snippet. Participants must label the text with emotions such as joy, sadness, fear, anger, surprise, and disgust. The goal is to determine which emotions are perceived to be present in the text.

%     \item \textbf{Track B: Emotion Intensity} \\
%     In this subtask, the objective is to predict the intensity of a given emotion in a text snippet. The intensity is categorized into ordinal classes: no emotion, low, moderate, and high degrees of emotion. This track focuses on quantifying how strongly an emotion is expressed in the text.

%     \item \textbf{Track C: Cross-lingual Emotion Detection} \\
%     This task challenges participants to predict perceived emotion labels for text instances in a different target language, using a labeled training set from one language. It emphasizes the ability to generalize emotion detection across different languages.
% \end{itemize}

This paper proposes EmoRAG, a Retrieval-Augmented Generation (RAG) system~\cite{rag_system} for the Subtask A, Multi-label Emotion Detection. However, its flexible design allows for seamless adaptation to the other subtasks, Emotion Intensity Prediction (Subtask B) and Cross-lingual Emotion Detection (Subtask C), with minimal modifications. This versatility makes EmoRAG a robust solution for the diverse challenges posed by SemEval 2025 Task 11.

% Developing generalizable solutions to these problems helps mitigate the risks of misusing generative language models (LMs) for malicious purposes~\cite{weidinger2022taxonomy} and improve human performance in identifying AI-produced content~\cite{gehrmann-etal-2019-gltr}.

%???
%This paper proposes EmoRAG, a novel method for multi-label emotion detection (Subtask A). 
%The boundary detection setup aligns with common user scenarios for applying generative LMs in practice, e.g., text continuation, creative writing, and story generation. 
% The standard approach to this task is training a linear classifier or a regression model over encoder representations~\cite{cutler2021automatic,dugan2023real}. 
%In contrast, EmoRAG leverages a pipeline of decoder and encoder models to detect machine-generated text, utilizing them sequentially. EmoRAG takes second out of 33 participating teams on the Subtask C leaderboard by achieving a Mean Absolute Error (MAE) of 15.94 on the official evaluation set. 

% The task of emotion recognition is complex due to the nuanced and subjective nature of emotions. 
% In this work, we participate in Track A of the shared task, which involves multi-label emotion detection. 
% The dataset includes text snippets in various languages, and the goal is to predict perceived emotions such as joy, sadness, fear, anger, surprise, and disgust. 
% This task emphasizes the perceived emotions of the speaker rather than the emotions evoked in the reader or the true emotions of the speaker.

\section{Background}

\begin{figure*}[htp!]
    \centering
    \includegraphics[width = 0.9\textwidth]{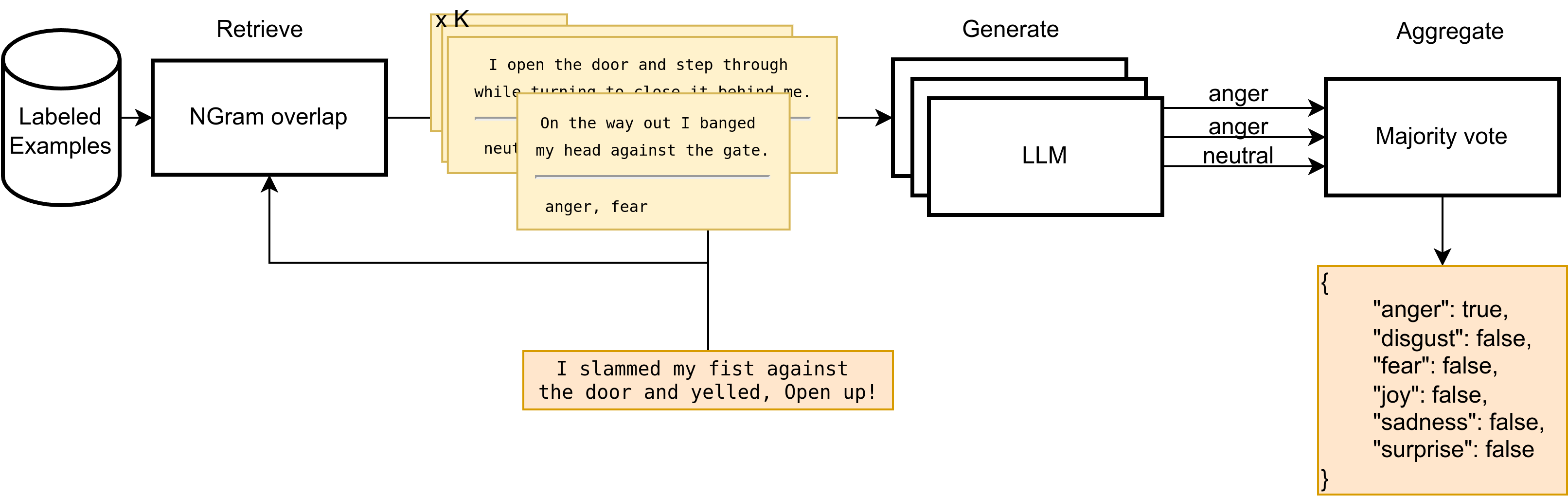}
    \caption{The EmoRAG pipeline involves a database, a retriever, a generator, and an aggregation model.}
    \label{fig:pipeline}
\end{figure*}

\paragraph{Related work}

A common approach to multi-label emotion classification involves fine-tuning a pre-trained transformer model with a linear classification head~\cite{kulkarni-etal-2021-pvg, kane-etal-2022-transformer}, often with minor architectural modifications to adapt to the specific task. Although such methods have shown strong performance in monolingual settings, emotion classification in a multilingual context presents additional challenges due to linguistic variability and cultural nuances in emotional expression~\cite{kadiyala-2024-cross}. To address these issues, we propose an alternative framework based on RAG. Unlike standard systems that rely solely on encoded representations, our method leverages the annotated training data as a retrieval corpus, enabling the model to draw on relevant emotional instances during inference, and thereby improve its robustness across languages and cultures.

\paragraph{Datasets}

The BRIGHTER dataset~\cite{muhammad2025brighterbridginggaphumanannotated} is a multilingual, multi-labeled collection of textual data annotated for emotion recognition in 28 languages. 
The dataset primarily addresses the disparity in emotion recognition resources, 
particularly for low-resource languages spoken in Africa, Asia, Eastern Europe, and Latin America.

The data is drawn from diverse sources, including social media posts, personal narratives, speeches, literary texts, 
and news articles, ensuring a broad representation of emotional expression across different cultural and linguistic contexts.

Each instance in the BRIGHTER dataset is manually curated and annotated by fluent speakers to capture six primary emotions: 
\emph{joy, sadness, anger, fear, surprise, disgust}, 
and a \emph{neutral} category. 
The annotations are multi-labeled, 
allowing each text snippet to be associated with multiple emotions.

The dataset encompasses both high-resource languages such as English and German and 
predominantly low-resource languages, including Hausa, Kinyarwanda, Emakhuwa, and isiZulu. 
The distribution of data sources varies across languages, 
with some relying on re-annotated sentiment datasets, 
human-written and machine-generated texts, and translated literary works. 
Notably, some datasets, such as Algerian Arabic, include translated excerpts from literary texts like \textit{La Grande Maison} by Mohammed Dib, 
whereas others, such as Hindi and Marathi, incorporate sentences generated by native speakers based on given prompts.

Some examples of the BRIGHTER dataset are shown below:
\begin{itemize}
    \item ``I can't believe this happened! I'm so excited and grateful!'' (Emotion labels: Joy, Surprise, Intensity: 3)
    \item ``Why do people always have to be so cruel? This is heartbreaking.'' (Emotion labels: Sadness, Anger, Intensity: 2)
    \item ``Walking through the dark alley gave me chills. I couldn't shake off the fear.'' (Emotion labels: Fear, Surprise, Intensity: 3)
\end{itemize}

Track A of the SemEval 2025 focuses on prediction emotions, ignoring intensity.

In addition, the EthioEmo dataset~\cite{emo}, introduced in a separate study, expands multilingual emotion recognition 
by incorporating four Ethiopian languages: Amharic, Afan Oromo, Somali, and Tigrinya. 
This extension further improves coverage for underrepresented languages, 
providing valuable benchmarks for evaluating large language models in multi-label emotion classification tasks.

It is important to note that the languages Zulu (zul), Xhosa (xho), Javanese (jav), and Indonesian (ind) were not part of the competition; 
therefore, the results for these languages are not presented in this paper.

%The BRIGHTER dataset, consisting of nearly 100,000 labeled instances, 
% is a significant step towards bridging the gap in multilingual emotion recognition 
% and fostering the development of inclusive NLP models for underrepresented languages. 
% It is publicly available for research purposes, promoting further advancements in cross-lingual emotion analysis 
% and related tasks.

\section{EmoRAG}\label{sec:emorag}

First, we overview the EmoRAG pipeline. Next, we detail the procedures for the database, retriever, generators, and aggregation model.

\paragraph{Overview} The EmoRAG pipeline consists of several components designed for emotion recognition:

\begin{enumerate}[label=(\alph*)]
    \item The database is created using labeled training examples.
    \item A retriever is used to fetch the top-$K$ most similar examples from the training data.
    \item The retrieved examples are used as few-shot prompts for the decoders, which are a collection of large language models (LLMs).
    \item An aggregation model combines the predictions from the generators to produce the final output.
\end{enumerate}

The overview of the EmoRAG pipeline is shown in Figure \ref{fig:pipeline}.

To make a prediction for a new entry in a known language, 
the system first uses the retriever to obtain the most similar examples from the database. 

Once the most similar examples are retrieved, these examples are used as few-shot prompts for the decoder models. Models utilize these prompts to predict the perceived emotions in the new text entry. Each decoder model produces a set of emotion predictions, which are then aggregated to form the final prediction.

The aggregation model combines the outputs from the different decoder models. 
This can be done using various strategies, such as majority voting or weighted averaging based on model performance metrics. The final output is a multi-label prediction indicating the perceived emotions present in the text.

\section{Experiments}

\begin{table*}[th!]
\centering
\footnotesize

\resizebox{.99\textwidth}{!}{

\begin{tabular}{@{}lcccccc@{}}
\toprule
\textbf{Language} & \textbf{Language Code} & \textbf{Best Model} & \textbf{Dev F1 Micro} & \textbf{Dev F1 Macro} & \textbf{Test F1 Micro} & \textbf{Test F1 Macro} \\ \midrule
Afrikaans & afr & majority\_vote\_by\_label\_f1 & 0.662 & 0.557 & 0.7153 & 0.667 \\
Amharic & amh & gpt-4o-mini & 0.637 & 0.503 & 0.6613 & 0.5578 \\
German & deu & gpt-4o-mini & 0.745 & 0.694 & 0.2694 & 0.2156 \\
English & eng & majority\_vote\_by\_label\_f1 & 0.821 & 0.818 & 0.8066 & 0.7885 \\
Spanish & esp & majority\_vote\_by\_label\_f1 & 0.813 & 0.809 & 0.8204 & 0.8174 \\
Hindi & hin & majority\_vote\_by\_label\_f1 & 0.842 & 0.849 & 0.8658 & 0.8661 \\
Marathi & mar & majority\_vote\_by\_label\_f1 & 0.943 & 0.947 & 0.8559 & 0.864 \\
Oromo & orm & gpt-4o-mini-ngram & 0.607 & 0.501 & 0.6023 & 0.4903 \\
Portuguese (Brazil) & ptbr & majority\_vote\_by\_label\_f1 & 0.766 & 0.645 & 0.4809 & 0.372 \\
Russian & rus & majority\_vote\_by\_label\_f1 & 0.880 & 0.880 & 0.8829 & 0.8794 \\
Somali & som & majority\_vote\_by\_label\_f1 & 0.519 & 0.477 & 0.5422 & 0.5082 \\
Sundanese & sun & gpt-4o-mini-ngram & 0.757 & 0.612 & 0.7256 & 0.5294 \\
Tatar & tat & majority\_vote\_by\_label\_f1 & 0.749 & 0.710 & 0.7884 & 0.7763 \\
Tigrinya & tir & majority\_vote\_by\_label\_f1 & 0.397 & 0.342 & 0.2597 & 0.2044 \\
Arabic (Algerian) & arq & majority\_vote\_by\_label\_f1 & 0.687 & 0.677 & 0.5464 & 0.5203 \\
Arabic (Moroccan) & ary & gpt-4o-mini-ngram & 0.576 & 0.512 & 0.4089 & 0.3701 \\
Chinese (Mandarin) & chn & gpt-4o-mini-ngram & 0.748 & 0.604 & 0.7416 & 0.6252 \\
Hausa & hau & majority\_vote\_by\_label\_f1 & 0.735 & 0.731 & 0.7039 & 0.6954 \\
Kinyarwanda & kin & gpt-4o-mini-ngram & 0.576 & 0.489 & 0.6167 & 0.5627 \\
Nigerian Pidgin & pcm & majority\_vote\_by\_label\_f1 & 0.638 & 0.591 & 0.6416 & 0.5993 \\
Portuguese (Mozambique) & ptmz & majority\_vote\_by\_label\_f1 & 0.565 & 0.558 & 0.535 & 0.4927 \\
Swahili & swa & majority\_vote\_by\_label\_f1 & 0.440 & 0.409 & 0.43 & 0.3856 \\
Swedish & swe & majority\_vote\_by\_label\_f1 & 0.736 & 0.582 & 0.6353 & 0.4926 \\
Ukrainian & ukr & majority\_vote\_by\_label\_f1 & 0.634 & 0.621 & 0.638 & 0.6161 \\
Emakhuwa & vmw & gpt-4o-mini & 0.300 & 0.211 & 0.2556 & 0.2157 \\
Yoruba & yor & majority\_vote\_by\_label\_f1 & 0.564 & 0.443 & 0.5257 & 0.3818 \\
Igbo & ibo & majority\_vote\_by\_label\_f1 & 0.614 & 0.550 & 0.6125 & 0.5379 \\
Romanian & ron & majority\_vote\_by\_label\_f1 & 0.794 & 0.774 & 0.773 & 0.7608 \\ \midrule
\textbf{Average} & & & & & \textbf{0.638} & \textbf{0.590} \\ \bottomrule
\end{tabular}
}
\caption{Test set performance metrics for each language using the best model according to the development dataset results.}
\label{tab:test_metrics}
\end{table*}

The main three components of EmoRAG are the retriever, the generator models, and the aggregation model.

\paragraph{Retrievers} We experimented with an n-gram based retriever and a sentence embedder-based retriever \textit{BGE-M3}~\footnote{\href{https://huggingface.co/BAAI/bge-m3}{\texttt{hf.co/meta-llamaBAAI/bge-m3}}}~\cite{bge-m3}. The n-gram based retriever is hypothesized to perform better for low-resource languages due to its reliance on surface-level text features. 
We have used the n-gram retriever from the 
\texttt{LangChain module}~\footnote{\href{https://python.langchain.com/docs/how_to/example_selectors_ngram}{\texttt{python.langchain.com}}}~\cite{Chase_LangChain_2022}. %For more details on the n-gram retriever, see the \url{https://python.langchain.com/docs/how_to/example_selectors_ngram/}.

The number of retrieved examples ($K$) is fixed to 30 for low-resource languages and 100 for high-resource languages.
The reason for this is that low-resource languages consume more tokens and thus more compute.

\paragraph{Decoder Models} The EmoRAG system employed four different LLMs: \textit{Llama-3.1-70B}~\footnote{\href{https://huggingface.co/meta-llama/Llama-3.1-70B-Instruct}{\texttt{hf.co/meta-llama/Llama-3.1-70B-Instruct}}}~\cite{grattafiori2024llama3herdmodels}, \textit{Qwen2.5-72B-Instruct}~\footnote{\href{https://huggingface.co/Qwen/Qwen2.5-72B-Instruct}{\texttt{hf.co/Qwen/Qwen2.5-72B-Instruct}}}~\cite{qwen2}, \textit{gpt-4o-mini-2024-07-18} (hereafter referred to as gpt-4o-mini), and \textit{gemma-2-27b-it}~\footnote{\href{https://huggingface.co/google/gemma-2-27b-it}{\texttt{hf.co/google/gemma-2-27b-it}}}~\cite{gemmateam2024gemma2improvingopen}. 
The LLM system prompt is in English and only specifies the language of the input text. We experimented with prompts in the target language for which predictions were made but found that English prompts yielded better results. We provide the whole system prompt in the appendix \ref{box:emotion_detection}.

\paragraph{Aggregation Strategies} We tested five aggregation strategies to combine predictions from different models: 

\begin{itemize}
    \item \textbf{Single Model}: Outputs the prediction of a fixed model, such as gpt-4o-mini.
    \item \textbf{Majority Vote}: Each label's prediction is the majority vote across all LLM predictions.
    \item \textbf{Macro/Micro Majority Vote}: Weighted averages of predictions from different LLMs, with weights based on macro/micro F1 scores on the dev data.
    \item \textbf{Label-F1 Majority Vote}: Weighted averages for each label, with weights based on macro/micro F1 scores for each label and model.
    \item \textbf{GPT-4o Aggregation}: Provides results from different models to gpt-4o-mini, along with few-shot examples, to aggregate a response.
\end{itemize}

\section{Results}

\begin{table*}[!th]
\begin{center}
    
    \footnotesize
    \resizebox{\textwidth}{!}{
    \begin{tabular}{@{}lccccccccc@{}}
    \toprule
    \textbf{Language} & \textbf{llama-3.1-70b} & \textbf{qwen2.5-70b} & \textbf{gpt-4o-mini} & \textbf{gpt-4o-mini-ngram} & \textbf{gemma29b} & \textbf{gemma29b\_ngram} & \textbf{majority\_vote} & \textbf{majority\_vote\_macro} & \textbf{majority\_vote\_by\_label\_f1} \\ \midrule
    amh & 0.534/0.448 & - & \textbf{0.637/0.503} & 0.633/0.493 & 0.609/0.488 & 0.582/0.474 & 0.659/0.535 & 0.659/0.535 & 0.655/0.539 \\
    arq & 0.584/0.575 & 0.623/0.597 & 0.613/0.596 & \textbf{0.663/0.655} & 0.614/0.589 & 0.578/0.531 & 0.645/0.615 & 0.653/0.665 & 0.687/0.677 \\
    ary & 0.542/0.490 & 0.540/0.485 & 0.552/0.499 & \textbf{0.576/0.512} & 0.575/0.521 & 0.584/0.484 & 0.607/0.526 & 0.526/0.599 & 0.616/0.540 \\
    afr & 0.560/0.444 & \textbf{0.629/0.527} & 0.662/0.567 & 0.646/0.572 & 0.584/0.481 & 0.484/0.398 & 0.601/0.494 & 0.546/0.646 & 0.662/0.557 \\
    chn & 0.676/0.603 & 0.589/0.570 & 0.698/0.579 & \textbf{0.748/0.604} & 0.693/0.572 & 0.709/0.543 & 0.749/0.642 & 0.652/0.757 & 0.759/0.659 \\
    deu & 0.745/0.588 & 0.521/0.499 & \textbf{0.745/0.694} & 0.738/0.662 & 0.632/0.559 & 0.659/0.593 & 0.738/0.659 & 0.672/0.741 & 0.752/0.695 \\
    eng & 0.735/0.726 & 0.779/0.775 & 0.807/0.803 & 0.770/0.781 & 0.769/0.759 & 0.720/0.723 & 0.801/0.808 & 0.823/0.820 & \textbf{0.821/0.818} \\
    esp & 0.751/0.744 & 0.788/0.778 & 0.793/0.785 & 0.799/0.793 & 0.778/0.772 & 0.782/0.778 & 0.786/0.778 & 0.807/0.812 & \textbf{0.813/0.809} \\
    hau & 0.610/0.602 & 0.607/0.598 & 0.669/0.662 & \textbf{0.696/0.687} & 0.682/0.676 & 0.698/0.689 & 0.735/0.728 & 0.734/0.738 & 0.735/0.731 \\
    hin & 0.780/0.791 & 0.707/0.728 & 0.805/0.803 & 0.811/0.812 & 0.796/0.799 & 0.798/0.806 & 0.838/0.842 & 0.833/0.830 & \textbf{0.842/0.849} \\
    ibo & 0.531/0.486 & 0.502/0.452 & 0.572/0.514 & 0.564/0.499 & 0.574/0.508 & 0.574/0.520 & 0.609/0.532 & 0.534/0.608 & \textbf{0.614/0.550} \\
    kin & 0.443/0.385 & 0.443/0.382 & 0.555/0.491 & \textbf{0.576/0.489} & 0.477/0.404 & 0.514/0.466 & 0.589/0.515 & 0.501/0.570 & 0.575/0.512 \\
    mar & 0.874/0.883 & 0.904/0.908 & 0.937/0.939 & 0.937/0.939 & 0.883/0.883 & 0.897/0.900 & 0.942/0.946 & 0.935/0.931 & \textbf{0.943/0.947} \\
    orm & 0.467/0.369 & 0.521/0.415 & 0.552/0.455 & \textbf{0.607/0.501} & 0.519/0.404 & 0.488/0.362 & 0.585/0.446 & 0.493/0.608 & 0.608/0.488 \\
    pcm & 0.532/0.508 & 0.573/0.535 & 0.599/0.542 & \textbf{0.628/0.573} & 0.608/0.572 & 0.585/0.548 & 0.621/0.574 & 0.590/0.633 & 0.638/0.591 \\
    ptbr & 0.686/0.547 & 0.662/0.569 & 0.731/0.633 & 0.707/0.603 & 0.726/0.617 & 0.710/0.525 & 0.766/0.626 & 0.658/0.760 & \textbf{0.766/0.645} \\
    ptmz & 0.454/0.456 & 0.539/0.532 & 0.515/0.484 & 0.478/0.443 & 0.521/0.486 & 0.494/0.445 & 0.565/0.558 & 0.543/0.552 & \textbf{0.565/0.558} \\
    ron & 0.758/0.749 & 0.745/0.726 & 0.756/0.741 & 0.778/0.763 & 0.745/0.719 & 0.754/0.724 & 0.773/0.751 & 0.771/0.790 & \textbf{0.794/0.774} \\
    rus & 0.835/0.836 & 0.861/0.857 & 0.839/0.833 & 0.812/0.806 & 0.841/0.834 & 0.824/0.817 & 0.879/0.877 & 0.881/0.883 & \textbf{0.880/0.880} \\
    som & 0.361/0.296 & 0.379/0.338 & 0.518/0.469 & 0.528/0.491 & 0.426/0.381 & 0.428/0.382 & 0.494/0.420 & 0.464/0.514 & \textbf{0.519/0.477} \\
    sun & 0.674/0.496 & 0.707/0.491 & 0.734/0.596 & 0.757/0.612 & 0.708/0.532 & 0.733/0.565 & 0.754/0.537 & 0.564/0.750 & \textbf{0.757/0.614} \\
    swa & 0.357/0.329 & 0.376/0.345 & 0.391/0.366 & 0.416/0.401 & 0.401/0.366 & 0.407/0.372 & 0.435/0.396 & 0.401/0.435 & \textbf{0.440/0.409} \\
    swe & 0.684/0.475 & 0.680/0.502 & 0.709/0.528 & 0.708/0.529 & 0.699/0.518 & 0.671/0.501 & 0.734/0.555 & 0.547/0.727 & \textbf{0.736/0.582} \\
    tat & 0.652/0.611 & 0.663/0.631 & 0.712/0.671 & 0.702/0.660 & 0.669/0.634 & 0.637/0.592 & 0.727/0.673 & 0.688/0.732 & \textbf{0.749/0.710} \\
    tir & - & - & 0.377/0.321 & 0.384/0.319 & - & - & 0.322/0.263 & 0.321/0.377 & \textbf{0.397/0.342} \\
    ukr & 0.521/0.512 & 0.601/0.579 & 0.581/0.567 & 0.550/0.537 & 0.587/0.553 & 0.535/0.469 & 0.622/0.611 & 0.621/0.625 & \textbf{0.634/0.621} \\
    vmw & 0.158/0.145 & 0.261/0.184 & 0.300/0.211 & 0.226/0.158 & 0.246/0.206 & 0.186/0.159 & 0.190/0.140 & 0.180/0.230 & \textbf{0.257/0.205} \\
    yor & 0.354/0.255 & 0.415/0.300 & 0.474/0.374 & 0.506/0.420 & 0.436/0.317 & 0.472/0.347 & 0.564/0.443 & 0.423/0.532 & \textbf{0.564/0.443} \\ \midrule
    \textbf{Average} & \textbf{0.563/0.515} & \textbf{0.590/0.556} & \textbf{0.631/0.590} & \textbf{0.641/0.601} & \textbf{0.617/0.576} & \textbf{0.607/0.566} & \textbf{0.661/0.617} & \textbf{0.646/0.634} & \textbf{0.678/0.634} \\ \bottomrule
    \end{tabular}
    }
\end{center}
\caption{Development set F1-micro/F1-macro scores for each language and model. The best model for each language is highlighted in bold.}
\label{tab:dev_metrics2}
\end{table*}

The results of our experiments are summarized in the Table~\ref{tab:test_metrics}. More detailed results are provided in Table~\ref{tab:dev_metrics2}, which includes the performance of each model separately on the development set. The EmoRAG system demonstrated strong performance across a wide range of languages, with the majority\_vote\_by\_label\_f1 aggregation strategy generally yielding the best results.

\paragraph{Performance Across Languages} The system achieved high F1-micro and F1-macro scores in high-resource languages such as English, Spanish, and Russian, with scores exceeding 0.80. In low-resource languages, the performance was more variable, but the system still achieved competitive results, particularly with the use of the n-gram retriever.

\paragraph{Best Models} For each language, the best-performing model and aggregation strategy were selected based on the development set results. The majority\_vote\_by\_label\_f1 strategy was often the best choice, indicating the effectiveness of leveraging label-specific F1 scores for aggregation.

\paragraph{General Observations} The experiments highlighted the importance of selecting appropriate retrievers and aggregation strategies based on the language and resource availability. The EmoRAG system's flexibility in adapting to different languages and tasks makes it a robust solution for multilingual emotion detection.

Overall, the EmoRAG system achieved an average test F1-micro score of 0.638 and an F1-macro score of 0.590 across all languages, demonstrating its effectiveness in the SemEval-2025 Task 11.

\section{Conclusion}
This paper presents the EmoRAG system submitted to SemEval-2025 Task 11. Our system achieved strong performance across multiple languages, demonstrating its effectiveness in multilingual emotion recognition. EmoRAG introduces a novel pipeline that integrates RAG with LLMs and an adaptive aggregation mechanism. The combination of diverse retrievers and model-specific aggregation strategies enables flexible and robust emotion detection, particularly for low-resource languages. We believe this approach holds significant potential for improving multilingual NLP tasks by leveraging retrieved examples to enhance model predictions. Our Future research will be focused on refining retrieval methods, exploring alternative RAG techniques, and investigating the use of smaller, more efficient models to improve scalability and accessibility across different computational environments.

\section*{Acknowledgments}
LM's and AS's work results from a research project implemented in the Basic Research Program at the National Research University Higher School of Economics (HSE University). 
We acknowledge the computational resources of HSE University's HPC facilities.

\section*{Limitations}
While EmoRAG demonstrates strong performance, it has certain limitations. The current dataset includes a limited set of emotions, making it unclear how the method would generalize to a broader range of emotions. Additionally, the approach may struggle with highly imbalanced class distributions or significant distribution shifts in the data. Future work will focus on addressing these challenges by testing on more diverse datasets and improving robustness to class imbalance and domain shifts.

\bibliography{custom}

\begin{thebibliography}{12}
\providecommand{\natexlab}[1]{#1}

\bibitem[{Belay et~al.(2025)Belay, Azime, Ayele, Sidorov, Klakow, Slusallek, Kolesnikova, and Yimam}]{emo}
Tadesse~Destaw Belay, Israel~Abebe Azime, Abinew~Ali Ayele, Grigori Sidorov, Dietrich Klakow, Philip Slusallek, Olga Kolesnikova, and Seid~Muhie Yimam. 2025.
\newblock \href {https://aclanthology.org/2025.coling-main.237/} {Evaluating the capabilities of large language models for multi-label emotion understanding}.
\newblock In \emph{Proceedings of the 31st International Conference on Computational Linguistics}, pages 3523--3540, Abu Dhabi, UAE. Association for Computational Linguistics.

\bibitem[{Chase(2022)}]{Chase_LangChain_2022}
Harrison Chase. 2022.
\newblock \href {https://github.com/langchain-ai/langchain} {{LangChain}}.

\bibitem[{Chen et~al.(2024)Chen, Xiao, Zhang, Luo, Lian, and Liu}]{bge-m3}
Jianlv Chen, Shitao Xiao, Peitian Zhang, Kun Luo, Defu Lian, and Zheng Liu. 2024.
\newblock \href {https://arxiv.org/abs/2402.03216} {Bge m3-embedding: Multi-lingual, multi-functionality, multi-granularity text embeddings through self-knowledge distillation}.
\newblock \emph{Preprint}, arXiv:2402.03216.

\bibitem[{Grattafiori et~al.(2024)Grattafiori, Dubey, Jauhri, Pandey, Kadian, Al-Dahle, Letman, Mathur, Schelten, Vaughan, Yang, Fan, Goyal, Hartshorn, Yang, Mitra, Sravankumar, Korenev, Hinsvark, Rao, Zhang, Rodriguez, Gregerson, Spataru, Roziere, Biron, Tang, Chern, Caucheteux, Nayak, Bi, Marra, McConnell, Keller, Touret, Wu, Wong, Ferrer, Nikolaidis, Allonsius, Song, Pintz, Livshits, Wyatt, Esiobu, Choudhary, Mahajan, Garcia-Olano, Perino, Hupkes, Lakomkin, AlBadawy, Lobanova, Dinan, Smith, Radenovic, Guzmán, Zhang, Synnaeve, Lee, Anderson, Thattai, Nail, Mialon, Pang, Cucurell, Nguyen, Korevaar, Xu, Touvron, Zarov, Ibarra, Kloumann, Misra, Evtimov, Zhang, Copet, Lee, Geffert, Vranes, Park, Mahadeokar, Shah, van~der Linde, Billock, Hong, Lee, Fu, Chi, Huang, Liu, Wang, Yu, Bitton, Spisak, Park, Rocca, Johnstun, Saxe, Jia, Alwala, Prasad, Upasani, Plawiak, Li, Heafield, Stone, El-Arini, Iyer, Malik, Chiu, Bhalla, Lakhotia, Rantala-Yeary, van~der Maaten, Chen, Tan, Jenkins, Martin, Madaan, Malo, Blecher,
  Landzaat, de~Oliveira, Muzzi, Pasupuleti, Singh, Paluri, Kardas, Tsimpoukelli, Oldham, Rita, Pavlova, Kambadur, Lewis, Si, Singh, Hassan, Goyal, Torabi, Bashlykov, Bogoychev, Chatterji, Zhang, Duchenne, Çelebi, Alrassy, Zhang, Li, Vasic, Weng, Bhargava, Dubal, Krishnan, Koura, Xu, He, Dong, Srinivasan, Ganapathy, Calderer, Cabral, Stojnic, Raileanu, Maheswari, Girdhar, Patel, Sauvestre, Polidoro, Sumbaly, Taylor, Silva, Hou, Wang, Hosseini, Chennabasappa, Singh, Bell, Kim, Edunov, Nie, Narang, Raparthy, Shen, Wan, Bhosale, Zhang, Vandenhende, Batra, Whitman, Sootla, Collot, Gururangan, Borodinsky, Herman, Fowler, Sheasha, Georgiou, Scialom, Speckbacher, Mihaylov, Xiao, Karn, Goswami, Gupta, Ramanathan, Kerkez, Gonguet, Do, Vogeti, Albiero, Petrovic, Chu, Xiong, Fu, Meers, Martinet, Wang, Wang, Tan, Xia, Xie, Jia, Wang, Goldschlag, Gaur, Babaei, Wen, Song, Zhang, Li, Mao, Coudert, Yan, Chen, Papakipos, Singh, Srivastava, Jain, Kelsey, Shajnfeld, Gangidi, Victoria, Goldstand, Menon, Sharma, Boesenberg,
  Baevski, Feinstein, Kallet, Sangani, Teo, Yunus, Lupu, Alvarado, Caples, Gu, Ho, Poulton, Ryan, Ramchandani, Dong, Franco, Goyal, Saraf, Chowdhury, Gabriel, Bharambe, Eisenman, Yazdan, James, Maurer, Leonhardi, Huang, Loyd, Paola, Paranjape, Liu, Wu, Ni, Hancock, Wasti, Spence, Stojkovic, Gamido, Montalvo, Parker, Burton, Mejia, Liu, Wang, Kim, Zhou, Hu, Chu, Cai, Tindal, Feichtenhofer, Gao, Civin, Beaty, Kreymer, Li, Adkins, Xu, Testuggine, David, Parikh, Liskovich, Foss, Wang, Le, Holland, Dowling, Jamil, Montgomery, Presani, Hahn, Wood, Le, Brinkman, Arcaute, Dunbar, Smothers, Sun, Kreuk, Tian, Kokkinos, Ozgenel, Caggioni, Kanayet, Seide, Florez, Schwarz, Badeer, Swee, Halpern, Herman, Sizov, Guangyi, Zhang, Lakshminarayanan, Inan, Shojanazeri, Zou, Wang, Zha, Habeeb, Rudolph, Suk, Aspegren, Goldman, Zhan, Damlaj, Molybog, Tufanov, Leontiadis, Veliche, Gat, Weissman, Geboski, Kohli, Lam, Asher, Gaya, Marcus, Tang, Chan, Zhen, Reizenstein, Teboul, Zhong, Jin, Yang, Cummings, Carvill, Shepard, McPhie,
  Torres, Ginsburg, Wang, Wu, U, Saxena, Khandelwal, Zand, Matosich, Veeraraghavan, Michelena, Li, Jagadeesh, Huang, Chawla, Huang, Chen, Garg, A, Silva, Bell, Zhang, Guo, Yu, Moshkovich, Wehrstedt, Khabsa, Avalani, Bhatt, Mankus, Hasson, Lennie, Reso, Groshev, Naumov, Lathi, Keneally, Liu, Seltzer, Valko, Restrepo, Patel, Vyatskov, Samvelyan, Clark, Macey, Wang, Hermoso, Metanat, Rastegari, Bansal, Santhanam, Parks, White, Bawa, Singhal, Egebo, Usunier, Mehta, Laptev, Dong, Cheng, Chernoguz, Hart, Salpekar, Kalinli, Kent, Parekh, Saab, Balaji, Rittner, Bontrager, Roux, Dollar, Zvyagina, Ratanchandani, Yuvraj, Liang, Alao, Rodriguez, Ayub, Murthy, Nayani, Mitra, Parthasarathy, Li, Hogan, Battey, Wang, Howes, Rinott, Mehta, Siby, Bondu, Datta, Chugh, Hunt, Dhillon, Sidorov, Pan, Mahajan, Verma, Yamamoto, Ramaswamy, Lindsay, Lindsay, Feng, Lin, Zha, Patil, Shankar, Zhang, Zhang, Wang, Agarwal, Sajuyigbe, Chintala, Max, Chen, Kehoe, Satterfield, Govindaprasad, Gupta, Deng, Cho, Virk, Subramanian, Choudhury,
  Goldman, Remez, Glaser, Best, Koehler, Robinson, Li, Zhang, Matthews, Chou, Shaked, Vontimitta, Ajayi, Montanez, Mohan, Kumar, Mangla, Ionescu, Poenaru, Mihailescu, Ivanov, Li, Wang, Jiang, Bouaziz, Constable, Tang, Wu, Wang, Wu, Gao, Kleinman, Chen, Hu, Jia, Qi, Li, Zhang, Zhang, Adi, Nam, Yu, Wang, Zhao, Hao, Qian, Li, He, Rait, DeVito, Rosnbrick, Wen, Yang, Zhao, and Ma}]{grattafiori2024llama3herdmodels}
Aaron Grattafiori, Abhimanyu Dubey, Abhinav Jauhri, Abhinav Pandey, Abhishek Kadian, Ahmad Al-Dahle, Aiesha Letman, Akhil Mathur, Alan Schelten, Alex Vaughan, Amy Yang, Angela Fan, Anirudh Goyal, Anthony Hartshorn, Aobo Yang, Archi Mitra, Archie Sravankumar, Artem Korenev, Arthur Hinsvark, Arun Rao, Aston Zhang, Aurelien Rodriguez, Austen Gregerson, Ava Spataru, Baptiste Roziere, Bethany Biron, Binh Tang, Bobbie Chern, Charlotte Caucheteux, Chaya Nayak, Chloe Bi, Chris Marra, Chris McConnell, Christian Keller, Christophe Touret, Chunyang Wu, Corinne Wong, Cristian~Canton Ferrer, Cyrus Nikolaidis, Damien Allonsius, Daniel Song, Danielle Pintz, Danny Livshits, Danny Wyatt, David Esiobu, Dhruv Choudhary, Dhruv Mahajan, Diego Garcia-Olano, Diego Perino, Dieuwke Hupkes, Egor Lakomkin, Ehab AlBadawy, Elina Lobanova, Emily Dinan, Eric~Michael Smith, Filip Radenovic, Francisco Guzmán, Frank Zhang, Gabriel Synnaeve, Gabrielle Lee, Georgia~Lewis Anderson, Govind Thattai, Graeme Nail, Gregoire Mialon, Guan Pang,
  Guillem Cucurell, Hailey Nguyen, Hannah Korevaar, Hu~Xu, Hugo Touvron, Iliyan Zarov, Imanol~Arrieta Ibarra, Isabel Kloumann, Ishan Misra, Ivan Evtimov, Jack Zhang, Jade Copet, Jaewon Lee, Jan Geffert, Jana Vranes, Jason Park, Jay Mahadeokar, Jeet Shah, Jelmer van~der Linde, Jennifer Billock, Jenny Hong, Jenya Lee, Jeremy Fu, Jianfeng Chi, Jianyu Huang, Jiawen Liu, Jie Wang, Jiecao Yu, Joanna Bitton, Joe Spisak, Jongsoo Park, Joseph Rocca, Joshua Johnstun, Joshua Saxe, Junteng Jia, Kalyan~Vasuden Alwala, Karthik Prasad, Kartikeya Upasani, Kate Plawiak, Ke~Li, Kenneth Heafield, Kevin Stone, Khalid El-Arini, Krithika Iyer, Kshitiz Malik, Kuenley Chiu, Kunal Bhalla, Kushal Lakhotia, Lauren Rantala-Yeary, Laurens van~der Maaten, Lawrence Chen, Liang Tan, Liz Jenkins, Louis Martin, Lovish Madaan, Lubo Malo, Lukas Blecher, Lukas Landzaat, Luke de~Oliveira, Madeline Muzzi, Mahesh Pasupuleti, Mannat Singh, Manohar Paluri, Marcin Kardas, Maria Tsimpoukelli, Mathew Oldham, Mathieu Rita, Maya Pavlova, Melanie Kambadur,
  Mike Lewis, Min Si, Mitesh~Kumar Singh, Mona Hassan, Naman Goyal, Narjes Torabi, Nikolay Bashlykov, Nikolay Bogoychev, Niladri Chatterji, Ning Zhang, Olivier Duchenne, Onur Çelebi, Patrick Alrassy, Pengchuan Zhang, Pengwei Li, Petar Vasic, Peter Weng, Prajjwal Bhargava, Pratik Dubal, Praveen Krishnan, Punit~Singh Koura, Puxin Xu, Qing He, Qingxiao Dong, Ragavan Srinivasan, Raj Ganapathy, Ramon Calderer, Ricardo~Silveira Cabral, Robert Stojnic, Roberta Raileanu, Rohan Maheswari, Rohit Girdhar, Rohit Patel, Romain Sauvestre, Ronnie Polidoro, Roshan Sumbaly, Ross Taylor, Ruan Silva, Rui Hou, Rui Wang, Saghar Hosseini, Sahana Chennabasappa, Sanjay Singh, Sean Bell, Seohyun~Sonia Kim, Sergey Edunov, Shaoliang Nie, Sharan Narang, Sharath Raparthy, Sheng Shen, Shengye Wan, Shruti Bhosale, Shun Zhang, Simon Vandenhende, Soumya Batra, Spencer Whitman, Sten Sootla, Stephane Collot, Suchin Gururangan, Sydney Borodinsky, Tamar Herman, Tara Fowler, Tarek Sheasha, Thomas Georgiou, Thomas Scialom, Tobias Speckbacher,
  Todor Mihaylov, Tong Xiao, Ujjwal Karn, Vedanuj Goswami, Vibhor Gupta, Vignesh Ramanathan, Viktor Kerkez, Vincent Gonguet, Virginie Do, Vish Vogeti, Vítor Albiero, Vladan Petrovic, Weiwei Chu, Wenhan Xiong, Wenyin Fu, Whitney Meers, Xavier Martinet, Xiaodong Wang, Xiaofang Wang, Xiaoqing~Ellen Tan, Xide Xia, Xinfeng Xie, Xuchao Jia, Xuewei Wang, Yaelle Goldschlag, Yashesh Gaur, Yasmine Babaei, Yi~Wen, Yiwen Song, Yuchen Zhang, Yue Li, Yuning Mao, Zacharie~Delpierre Coudert, Zheng Yan, Zhengxing Chen, Zoe Papakipos, Aaditya Singh, Aayushi Srivastava, Abha Jain, Adam Kelsey, Adam Shajnfeld, Adithya Gangidi, Adolfo Victoria, Ahuva Goldstand, Ajay Menon, Ajay Sharma, Alex Boesenberg, Alexei Baevski, Allie Feinstein, Amanda Kallet, Amit Sangani, Amos Teo, Anam Yunus, Andrei Lupu, Andres Alvarado, Andrew Caples, Andrew Gu, Andrew Ho, Andrew Poulton, Andrew Ryan, Ankit Ramchandani, Annie Dong, Annie Franco, Anuj Goyal, Aparajita Saraf, Arkabandhu Chowdhury, Ashley Gabriel, Ashwin Bharambe, Assaf Eisenman, Azadeh
  Yazdan, Beau James, Ben Maurer, Benjamin Leonhardi, Bernie Huang, Beth Loyd, Beto~De Paola, Bhargavi Paranjape, Bing Liu, Bo~Wu, Boyu Ni, Braden Hancock, Bram Wasti, Brandon Spence, Brani Stojkovic, Brian Gamido, Britt Montalvo, Carl Parker, Carly Burton, Catalina Mejia, Ce~Liu, Changhan Wang, Changkyu Kim, Chao Zhou, Chester Hu, Ching-Hsiang Chu, Chris Cai, Chris Tindal, Christoph Feichtenhofer, Cynthia Gao, Damon Civin, Dana Beaty, Daniel Kreymer, Daniel Li, David Adkins, David Xu, Davide Testuggine, Delia David, Devi Parikh, Diana Liskovich, Didem Foss, Dingkang Wang, Duc Le, Dustin Holland, Edward Dowling, Eissa Jamil, Elaine Montgomery, Eleonora Presani, Emily Hahn, Emily Wood, Eric-Tuan Le, Erik Brinkman, Esteban Arcaute, Evan Dunbar, Evan Smothers, Fei Sun, Felix Kreuk, Feng Tian, Filippos Kokkinos, Firat Ozgenel, Francesco Caggioni, Frank Kanayet, Frank Seide, Gabriela~Medina Florez, Gabriella Schwarz, Gada Badeer, Georgia Swee, Gil Halpern, Grant Herman, Grigory Sizov, Guangyi, Zhang, Guna
  Lakshminarayanan, Hakan Inan, Hamid Shojanazeri, Han Zou, Hannah Wang, Hanwen Zha, Haroun Habeeb, Harrison Rudolph, Helen Suk, Henry Aspegren, Hunter Goldman, Hongyuan Zhan, Ibrahim Damlaj, Igor Molybog, Igor Tufanov, Ilias Leontiadis, Irina-Elena Veliche, Itai Gat, Jake Weissman, James Geboski, James Kohli, Janice Lam, Japhet Asher, Jean-Baptiste Gaya, Jeff Marcus, Jeff Tang, Jennifer Chan, Jenny Zhen, Jeremy Reizenstein, Jeremy Teboul, Jessica Zhong, Jian Jin, Jingyi Yang, Joe Cummings, Jon Carvill, Jon Shepard, Jonathan McPhie, Jonathan Torres, Josh Ginsburg, Junjie Wang, Kai Wu, Kam~Hou U, Karan Saxena, Kartikay Khandelwal, Katayoun Zand, Kathy Matosich, Kaushik Veeraraghavan, Kelly Michelena, Keqian Li, Kiran Jagadeesh, Kun Huang, Kunal Chawla, Kyle Huang, Lailin Chen, Lakshya Garg, Lavender A, Leandro Silva, Lee Bell, Lei Zhang, Liangpeng Guo, Licheng Yu, Liron Moshkovich, Luca Wehrstedt, Madian Khabsa, Manav Avalani, Manish Bhatt, Martynas Mankus, Matan Hasson, Matthew Lennie, Matthias Reso, Maxim
  Groshev, Maxim Naumov, Maya Lathi, Meghan Keneally, Miao Liu, Michael~L. Seltzer, Michal Valko, Michelle Restrepo, Mihir Patel, Mik Vyatskov, Mikayel Samvelyan, Mike Clark, Mike Macey, Mike Wang, Miquel~Jubert Hermoso, Mo~Metanat, Mohammad Rastegari, Munish Bansal, Nandhini Santhanam, Natascha Parks, Natasha White, Navyata Bawa, Nayan Singhal, Nick Egebo, Nicolas Usunier, Nikhil Mehta, Nikolay~Pavlovich Laptev, Ning Dong, Norman Cheng, Oleg Chernoguz, Olivia Hart, Omkar Salpekar, Ozlem Kalinli, Parkin Kent, Parth Parekh, Paul Saab, Pavan Balaji, Pedro Rittner, Philip Bontrager, Pierre Roux, Piotr Dollar, Polina Zvyagina, Prashant Ratanchandani, Pritish Yuvraj, Qian Liang, Rachad Alao, Rachel Rodriguez, Rafi Ayub, Raghotham Murthy, Raghu Nayani, Rahul Mitra, Rangaprabhu Parthasarathy, Raymond Li, Rebekkah Hogan, Robin Battey, Rocky Wang, Russ Howes, Ruty Rinott, Sachin Mehta, Sachin Siby, Sai~Jayesh Bondu, Samyak Datta, Sara Chugh, Sara Hunt, Sargun Dhillon, Sasha Sidorov, Satadru Pan, Saurabh Mahajan,
  Saurabh Verma, Seiji Yamamoto, Sharadh Ramaswamy, Shaun Lindsay, Shaun Lindsay, Sheng Feng, Shenghao Lin, Shengxin~Cindy Zha, Shishir Patil, Shiva Shankar, Shuqiang Zhang, Shuqiang Zhang, Sinong Wang, Sneha Agarwal, Soji Sajuyigbe, Soumith Chintala, Stephanie Max, Stephen Chen, Steve Kehoe, Steve Satterfield, Sudarshan Govindaprasad, Sumit Gupta, Summer Deng, Sungmin Cho, Sunny Virk, Suraj Subramanian, Sy~Choudhury, Sydney Goldman, Tal Remez, Tamar Glaser, Tamara Best, Thilo Koehler, Thomas Robinson, Tianhe Li, Tianjun Zhang, Tim Matthews, Timothy Chou, Tzook Shaked, Varun Vontimitta, Victoria Ajayi, Victoria Montanez, Vijai Mohan, Vinay~Satish Kumar, Vishal Mangla, Vlad Ionescu, Vlad Poenaru, Vlad~Tiberiu Mihailescu, Vladimir Ivanov, Wei Li, Wenchen Wang, Wenwen Jiang, Wes Bouaziz, Will Constable, Xiaocheng Tang, Xiaojian Wu, Xiaolan Wang, Xilun Wu, Xinbo Gao, Yaniv Kleinman, Yanjun Chen, Ye~Hu, Ye~Jia, Ye~Qi, Yenda Li, Yilin Zhang, Ying Zhang, Yossi Adi, Youngjin Nam, Yu, Wang, Yu~Zhao, Yuchen Hao, Yundi
  Qian, Yunlu Li, Yuzi He, Zach Rait, Zachary DeVito, Zef Rosnbrick, Zhaoduo Wen, Zhenyu Yang, Zhiwei Zhao, and Zhiyu Ma. 2024.
\newblock \href {https://arxiv.org/abs/2407.21783} {The llama 3 herd of models}.
\newblock \emph{Preprint}, arXiv:2407.21783.

\bibitem[{Kadiyala(2024)}]{kadiyala-2024-cross}
Ram Mohan~Rao Kadiyala. 2024.
\newblock \href {https://doi.org/10.18653/v1/2024.wassa-1.44} {Cross-lingual emotion detection through large language models}.
\newblock In \emph{Proceedings of the 14th Workshop on Computational Approaches to Subjectivity, Sentiment, {\&} Social Media Analysis}, pages 464--469, Bangkok, Thailand. Association for Computational Linguistics.

\bibitem[{Kane et~al.(2022)Kane, Patankar, Khose, and Kirtane}]{kane-etal-2022-transformer}
Aditya Kane, Shantanu Patankar, Sahil Khose, and Neeraja Kirtane. 2022.
\newblock \href {https://doi.org/10.18653/v1/2022.wassa-1.25} {Transformer based ensemble for emotion detection}.
\newblock In \emph{Proceedings of the 12th Workshop on Computational Approaches to Subjectivity, Sentiment {\&} Social Media Analysis}, pages 250--254, Dublin, Ireland. Association for Computational Linguistics.

\bibitem[{Kulkarni et~al.(2021)Kulkarni, Somwase, Rajput, and Marathe}]{kulkarni-etal-2021-pvg}
Atharva Kulkarni, Sunanda Somwase, Shivam Rajput, and Manisha Marathe. 2021.
\newblock \href {https://aclanthology.org/2021.wassa-1.11/} {{PVG} at {WASSA} 2021: A multi-input, multi-task, transformer-based architecture for empathy and distress prediction}.
\newblock In \emph{Proceedings of the Eleventh Workshop on Computational Approaches to Subjectivity, Sentiment and Social Media Analysis}, pages 105--111, Online. Association for Computational Linguistics.

\bibitem[{Lewis et~al.(2020)Lewis, Perez, Piktus, Petroni, Karpukhin, Goyal, K\"{u}ttler, Lewis, Yih, Rockt\"{a}schel, Riedel, and Kiela}]{rag_system}
Patrick Lewis, Ethan Perez, Aleksandra Piktus, Fabio Petroni, Vladimir Karpukhin, Naman Goyal, Heinrich K\"{u}ttler, Mike Lewis, Wen-tau Yih, Tim Rockt\"{a}schel, Sebastian Riedel, and Douwe Kiela. 2020.
\newblock Retrieval-augmented generation for knowledge-intensive nlp tasks.
\newblock In \emph{Proceedings of the 34th International Conference on Neural Information Processing Systems}, NIPS '20, Red Hook, NY, USA. Curran Associates Inc.

\bibitem[{Muhammad et~al.(2025{\natexlab{a}})Muhammad, Ousidhoum, Abdulmumin, Wahle, Ruas, Beloucif, de~Kock, Surange, Teodorescu, Ahmad, Adelani, Aji, Ali, Alimova, Araujo, Babakov, Baes, Bucur, Bukula, Cao, Cardenas, Chevi, Chukwuneke, Ciobotaru, Dementieva, Gadanya, Geislinger, Gipp, Hourrane, Ignat, Lawan, Mabuya, Mahendra, Marivate, Piper, Panchenko, Porto~Ferreira, Protasov, Rutunda, Shrivastava, Udrea, Wanzare, Wu, Wunderlich, Zhafran, Zhang, Zhou, and Mohammad}]{muhammad2025brighterbridginggaphumanannotated}
Shamsuddeen~Hassan Muhammad, Nedjma Ousidhoum, Idris Abdulmumin, Jan~Philip Wahle, Terry Ruas, Meriem Beloucif, Christine de~Kock, Nirmal Surange, Daniela Teodorescu, Ibrahim~Said Ahmad, David~Ifeoluwa Adelani, Alham~Fikri Aji, Felermino D. M.~A. Ali, Ilseyar Alimova, Vladimir Araujo, Nikolay Babakov, Naomi Baes, Ana-Maria Bucur, Andiswa Bukula, Guanqun Cao, Rodrigo~Tufino Cardenas, Rendi Chevi, Chiamaka~Ijeoma Chukwuneke, Alexandra Ciobotaru, Daryna Dementieva, Murja~Sani Gadanya, Robert Geislinger, Bela Gipp, Oumaima Hourrane, Oana Ignat, Falalu~Ibrahim Lawan, Rooweither Mabuya, Rahmad Mahendra, Vukosi Marivate, Andrew Piper, Alexander Panchenko, Charles~Henrique Porto~Ferreira, Vitaly Protasov, Samuel Rutunda, Manish Shrivastava, Aura~Cristina Udrea, Lilian Diana~Awuor Wanzare, Sophie Wu, Florian~Valentin Wunderlich, Hanif~Muhammad Zhafran, Tianhui Zhang, Yi~Zhou, and Saif~M. Mohammad. 2025{\natexlab{a}}.
\newblock Brighter: Bridging the gap in human-annotated textual emotion recognition datasets for 28 languages.
\newblock \emph{arXiv preprint arXiv:2502.11926}.

\bibitem[{Muhammad et~al.(2025{\natexlab{b}})Muhammad, Ousidhoum, Abdulmumin, Yimam, Wahle, Ruas, Beloucif, De~Kock, Belay, Ahmad, Surange, Teodorescu, Adelani, Aji, Ali, Araujo, Ayele, Ignat, Panchenko, Zhou, and Mohammad}]{muhammad-etal-2025-semeval1}
Shamsuddeen~Hassan Muhammad, Nedjma Ousidhoum, Idris Abdulmumin, Seid~Muhie Yimam, Jan~Philip Wahle, Terry Ruas, Meriem Beloucif, Christine De~Kock, Tadesse~Destaw Belay, Ibrahim~Said Ahmad, Nirmal Surange, Daniela Teodorescu, David~Ifeoluwa Adelani, Alham~Fikri Aji, Felermino Ali, Vladimir Araujo, Abinew~Ali Ayele, Oana Ignat, Alexander Panchenko, Yi~Zhou, and Saif~M. Mohammad. 2025{\natexlab{b}}.
\newblock {S}em{E}val task 11: Bridging the gap in text-based emotion detection.
\newblock In \emph{Proceedings of the 19th International Workshop on Semantic Evaluation (SemEval-2025)}, Vienna, Austria. Association for Computational Linguistics.

\bibitem[{Team et~al.(2024)Team, Riviere, Pathak, Sessa, Hardin, Bhupatiraju, Hussenot, Mesnard, Shahriari, Ramé, Ferret, Liu, Tafti, Friesen, Casbon, Ramos, Kumar, Lan, Jerome, Tsitsulin, Vieillard, Stanczyk, Girgin, Momchev, Hoffman, Thakoor, Grill, Neyshabur, Bachem, Walton, Severyn, Parrish, Ahmad, Hutchison, Abdagic, Carl, Shen, Brock, Coenen, Laforge, Paterson, Bastian, Piot, Wu, Royal, Chen, Kumar, Perry, Welty, Choquette-Choo, Sinopalnikov, Weinberger, Vijaykumar, Rogozińska, Herbison, Bandy, Wang, Noland, Moreira, Senter, Eltyshev, Visin, Rasskin, Wei, Cameron, Martins, Hashemi, Klimczak-Plucińska, Batra, Dhand, Nardini, Mein, Zhou, Svensson, Stanway, Chan, Zhou, Carrasqueira, Iljazi, Becker, Fernandez, van Amersfoort, Gordon, Lipschultz, Newlan, yeong Ji, Mohamed, Badola, Black, Millican, McDonell, Nguyen, Sodhia, Greene, Sjoesund, Usui, Sifre, Heuermann, Lago, McNealus, Soares, Kilpatrick, Dixon, Martins, Reid, Singh, Iverson, Görner, Velloso, Wirth, Davidow, Miller, Rahtz, Watson, Risdal,
  Kazemi, Moynihan, Zhang, Kahng, Park, Rahman, Khatwani, Dao, Bardoliwalla, Devanathan, Dumai, Chauhan, Wahltinez, Botarda, Barnes, Barham, Michel, Jin, Georgiev, Culliton, Kuppala, Comanescu, Merhej, Jana, Rokni, Agarwal, Mullins, Saadat, Carthy, Cogan, Perrin, Arnold, Krause, Dai, Garg, Sheth, Ronstrom, Chan, Jordan, Yu, Eccles, Hennigan, Kocisky, Doshi, Jain, Yadav, Meshram, Dharmadhikari, Barkley, Wei, Ye, Han, Kwon, Xu, Shen, Gong, Wei, Cotruta, Kirk, Rao, Giang, Peran, Warkentin, Collins, Barral, Ghahramani, Hadsell, Sculley, Banks, Dragan, Petrov, Vinyals, Dean, Hassabis, Kavukcuoglu, Farabet, Buchatskaya, Borgeaud, Fiedel, Joulin, Kenealy, Dadashi, and Andreev}]{gemmateam2024gemma2improvingopen}
Gemma Team, Morgane Riviere, Shreya Pathak, Pier~Giuseppe Sessa, Cassidy Hardin, Surya Bhupatiraju, Léonard Hussenot, Thomas Mesnard, Bobak Shahriari, Alexandre Ramé, Johan Ferret, Peter Liu, Pouya Tafti, Abe Friesen, Michelle Casbon, Sabela Ramos, Ravin Kumar, Charline~Le Lan, Sammy Jerome, Anton Tsitsulin, Nino Vieillard, Piotr Stanczyk, Sertan Girgin, Nikola Momchev, Matt Hoffman, Shantanu Thakoor, Jean-Bastien Grill, Behnam Neyshabur, Olivier Bachem, Alanna Walton, Aliaksei Severyn, Alicia Parrish, Aliya Ahmad, Allen Hutchison, Alvin Abdagic, Amanda Carl, Amy Shen, Andy Brock, Andy Coenen, Anthony Laforge, Antonia Paterson, Ben Bastian, Bilal Piot, Bo~Wu, Brandon Royal, Charlie Chen, Chintu Kumar, Chris Perry, Chris Welty, Christopher~A. Choquette-Choo, Danila Sinopalnikov, David Weinberger, Dimple Vijaykumar, Dominika Rogozińska, Dustin Herbison, Elisa Bandy, Emma Wang, Eric Noland, Erica Moreira, Evan Senter, Evgenii Eltyshev, Francesco Visin, Gabriel Rasskin, Gary Wei, Glenn Cameron, Gus Martins,
  Hadi Hashemi, Hanna Klimczak-Plucińska, Harleen Batra, Harsh Dhand, Ivan Nardini, Jacinda Mein, Jack Zhou, James Svensson, Jeff Stanway, Jetha Chan, Jin~Peng Zhou, Joana Carrasqueira, Joana Iljazi, Jocelyn Becker, Joe Fernandez, Joost van Amersfoort, Josh Gordon, Josh Lipschultz, Josh Newlan, Ju~yeong Ji, Kareem Mohamed, Kartikeya Badola, Kat Black, Katie Millican, Keelin McDonell, Kelvin Nguyen, Kiranbir Sodhia, Kish Greene, Lars~Lowe Sjoesund, Lauren Usui, Laurent Sifre, Lena Heuermann, Leticia Lago, Lilly McNealus, Livio~Baldini Soares, Logan Kilpatrick, Lucas Dixon, Luciano Martins, Machel Reid, Manvinder Singh, Mark Iverson, Martin Görner, Mat Velloso, Mateo Wirth, Matt Davidow, Matt Miller, Matthew Rahtz, Matthew Watson, Meg Risdal, Mehran Kazemi, Michael Moynihan, Ming Zhang, Minsuk Kahng, Minwoo Park, Mofi Rahman, Mohit Khatwani, Natalie Dao, Nenshad Bardoliwalla, Nesh Devanathan, Neta Dumai, Nilay Chauhan, Oscar Wahltinez, Pankil Botarda, Parker Barnes, Paul Barham, Paul Michel, Pengchong Jin,
  Petko Georgiev, Phil Culliton, Pradeep Kuppala, Ramona Comanescu, Ramona Merhej, Reena Jana, Reza~Ardeshir Rokni, Rishabh Agarwal, Ryan Mullins, Samaneh Saadat, Sara~Mc Carthy, Sarah Cogan, Sarah Perrin, Sébastien M.~R. Arnold, Sebastian Krause, Shengyang Dai, Shruti Garg, Shruti Sheth, Sue Ronstrom, Susan Chan, Timothy Jordan, Ting Yu, Tom Eccles, Tom Hennigan, Tomas Kocisky, Tulsee Doshi, Vihan Jain, Vikas Yadav, Vilobh Meshram, Vishal Dharmadhikari, Warren Barkley, Wei Wei, Wenming Ye, Woohyun Han, Woosuk Kwon, Xiang Xu, Zhe Shen, Zhitao Gong, Zichuan Wei, Victor Cotruta, Phoebe Kirk, Anand Rao, Minh Giang, Ludovic Peran, Tris Warkentin, Eli Collins, Joelle Barral, Zoubin Ghahramani, Raia Hadsell, D.~Sculley, Jeanine Banks, Anca Dragan, Slav Petrov, Oriol Vinyals, Jeff Dean, Demis Hassabis, Koray Kavukcuoglu, Clement Farabet, Elena Buchatskaya, Sebastian Borgeaud, Noah Fiedel, Armand Joulin, Kathleen Kenealy, Robert Dadashi, and Alek Andreev. 2024.
\newblock \href {https://arxiv.org/abs/2408.00118} {Gemma 2: Improving open language models at a practical size}.
\newblock \emph{Preprint}, arXiv:2408.00118.

\bibitem[{Yang et~al.(2024)Yang, Yang, Hui, Zheng, Yu, Zhou, Li, Li, Liu, Huang, Dong, Wei, Lin, Tang, Wang, Yang, Tu, Zhang, Ma, Xu, Zhou, Bai, He, Lin, Dang, Lu, Chen, Yang, Li, Xue, Ni, Zhang, Wang, Peng, Men, Gao, Lin, Wang, Bai, Tan, Zhu, Li, Liu, Ge, Deng, Zhou, Ren, Zhang, Wei, Ren, Fan, Yao, Zhang, Wan, Chu, Liu, Cui, Zhang, and Fan}]{qwen2}
An~Yang, Baosong Yang, Binyuan Hui, Bo~Zheng, Bowen Yu, Chang Zhou, Chengpeng Li, Chengyuan Li, Dayiheng Liu, Fei Huang, Guanting Dong, Haoran Wei, Huan Lin, Jialong Tang, Jialin Wang, Jian Yang, Jianhong Tu, Jianwei Zhang, Jianxin Ma, Jin Xu, Jingren Zhou, Jinze Bai, Jinzheng He, Junyang Lin, Kai Dang, Keming Lu, Keqin Chen, Kexin Yang, Mei Li, Mingfeng Xue, Na~Ni, Pei Zhang, Peng Wang, Ru~Peng, Rui Men, Ruize Gao, Runji Lin, Shijie Wang, Shuai Bai, Sinan Tan, Tianhang Zhu, Tianhao Li, Tianyu Liu, Wenbin Ge, Xiaodong Deng, Xiaohuan Zhou, Xingzhang Ren, Xinyu Zhang, Xipin Wei, Xuancheng Ren, Yang Fan, Yang Yao, Yichang Zhang, Yu~Wan, Yunfei Chu, Yuqiong Liu, Zeyu Cui, Zhenru Zhang, and Zhihao Fan. 2024.
\newblock Qwen2 technical report.
\newblock \emph{arXiv preprint arXiv:2407.10671}.

\end{thebibliography}

\appendix

\section{LLM Prompt for Emotion Detection} 
\label{appendix:llm_prompt}

\lstset{
  basicstyle=\small\ttfamily,
  breaklines=true,
  breakatwhitespace=false,
  breakindent=0pt,
  literate={-}{{-}}1, % ensures hyphens are treated properly
}

The following prompt is used to instruct the language model for perceived emotion detection:
\DefineVerbatimEnvironment{MyVerbatim}{Verbatim}{breaklines=true, breaksymbolleft==\phantom{}, fontsize=\small}

\begin{tcolorbox}[colback=orange!5, colframe=orange!90!black, fonttitle=\bfseries, title=Emotion Detection Prompt, boxrule=1pt, arc=4pt, auto outer arc, boxsep=5pt, left=5pt, right=5pt, top=5pt, bottom=5pt, halign=justify]
\label{box:emotion_detection}
\begin{MyVerbatim}
You are an expert at detecting emotions in text. The texts are given in {language} language.
Please classify the text into one of the following categories:
Anger, Fear, Joy, Sadness, Surprise, Disgust
Your response should be a JSON object with the following format:
{
    "anger": bool,
    "fear": bool,
    "joy": bool,
    "sadness": bool,
    "surprise": bool,
    "disgust": bool
}
Do not give explanations. Just return the JSON object.
\end{MyVerbatim}
\end{tcolorbox}

\end{document}